\def\paperID{10556}
\def\confName{ICCV}
\def\confYear{2025}

\newif\ifreview 
\newif\ifarxiv 
\newif\ifcamera \newcommand{\cameraready}{\cameratrue}
\newif\ifrebuttal 

\cameraready

\pdfoutput=1
\documentclass[10pt,twocolumn,letterpaper]{article}
\ifreview \usepackage[review]{cvpr} \fi
\ifarxiv \usepackage[pagenumbers]{cvpr} \fi
\ifrebuttal \usepackage[rebuttal]{cvpr} \fi
\ifcamera \usepackage{cvpr} \fi

\usepackage{graphicx}	
\usepackage{amsmath}
\usepackage{bm}
\usepackage{amssymb}	
\usepackage{booktabs}
\usepackage{times}
\usepackage{microtype}
\usepackage{epsfig}
\usepackage{caption}
\usepackage{pifont} 
\usepackage{float}
\usepackage{placeins}
\usepackage{color, colortbl}
\usepackage{stfloats}
\usepackage{enumitem}
\usepackage{tabularx}
\usepackage{xstring}
\usepackage{multirow}
\usepackage{xspace}
\usepackage{url}
\usepackage{subcaption}
\usepackage{xcolor}
\usepackage{indentfirst}
\usepackage{array}
\usepackage{booktabs}

\usepackage[hang,flushmargin]{footmisc}
\usepackage{booktabs}   
\usepackage{pifont}     
\usepackage{arydshln}   
\ifcamera \usepackage[accsupp]{axessibility} \fi

% ...

\ifarxiv  \fi

\newcommand{\R}[1]{{%
    \textbf{%
        \ifstrequal{#1}{1}{\textcolor{red}{R#1}}{%
        \ifstrequal{#1}{2}{\textcolor{blue}{R#1}}{%
        \ifstrequal{#1}{3}{\textcolor{magenta}{R#1}}{%
        \ifstrequal{#1}{4}{\textcolor{teal}{R#1}}{%
                           \textcolor{cyan}{R#1}%
        }}}}%
    }%
}}

\usepackage{xr-hyper}

\makeatletter
\newcommand*{\addFileDependency}[1]{
  \typeout{(#1)}
  \@addtofilelist{#1}
  \IfFileExists{#1}{}{\typeout{No file #1.}}
}

\makeatother
\newcommand*{\myexternaldocument}[1]{
    \externaldocument{#1}
    \addFileDependency{#1.tex}
    \addFileDependency{#1.aux}
}

\definecolor{cvprblue}{rgb}{0.21,0.49,0.74}
\usepackage[pagebackref,breaklinks,colorlinks,allcolors=cvprblue]{hyperref}
\usepackage[capitalize]{cleveref}
\crefname{section}{Sec.}{Secs.}
\crefname{table}{Table}{Tables}
\crefname{figure}{Fig.}{Figs.}

\ifarxiv \crefname{appendix}{App.}{Apps.}
\else \crefname{appendix}{Suppl.}{Suppls.} \fi

\unless\ifarxiv \myexternaldocument{_supplementary} \fi

\begin{document}
\title{Information-Bottleneck Driven Binary Neural Network for Change Detection}

\author{
Kaijie Yin$^{1}$\hspace{0.02in}
Zhiyuan Zhang$^{2}$ \hspace{0.02in}
Shu Kong$^{1}$ \hspace{0.02in}
Tian Gao$^{1,3}$ \hspace{0.02in}
Chengzhong Xu$^{1}$ \hspace{0.02in}
Hui Kong$^{1}$ \thanks{Corresponding author}
\vspace{0.1in}\\
$^{1}$University of Macau \quad
$^{2}$Singapore Management University 
\\
$^{3}$Nanjing University of Science and Technology
}

\maketitle

\begin{abstract}

In this paper, we propose Binarized Change Detection (\textbf{BiCD}), the first binary neural network (BNN) designed specifically for change detection. Conventional network binarization approaches, which directly quantize both weights and activations in change detection models, severely limit the network's ability to represent input data and distinguish between changed and unchanged regions. This results in significantly lower detection accuracy compared to real-valued networks.
To overcome these challenges, BiCD enhances both the representational power and feature separability of BNNs, improving detection performance. Specifically, we introduce an auxiliary objective based on the Information Bottleneck (IB) principle, guiding the encoder to retain essential input information while promoting better feature discrimination. Since directly computing mutual information under the IB principle is intractable, we design a compact, learnable auxiliary module as an approximation target, leading to a simple yet effective optimization strategy that minimizes both reconstruction loss and standard change detection loss.
Extensive experiments on street-view and remote sensing datasets demonstrate that BiCD establishes a new benchmark for BNN-based change detection, achieving state-of-the-art performance in this domain.

\end{abstract}
\section{Introduction}
\label{sec:intro}


\begin{figure}
\centering
\includegraphics[width=3.2in, keepaspectratio]{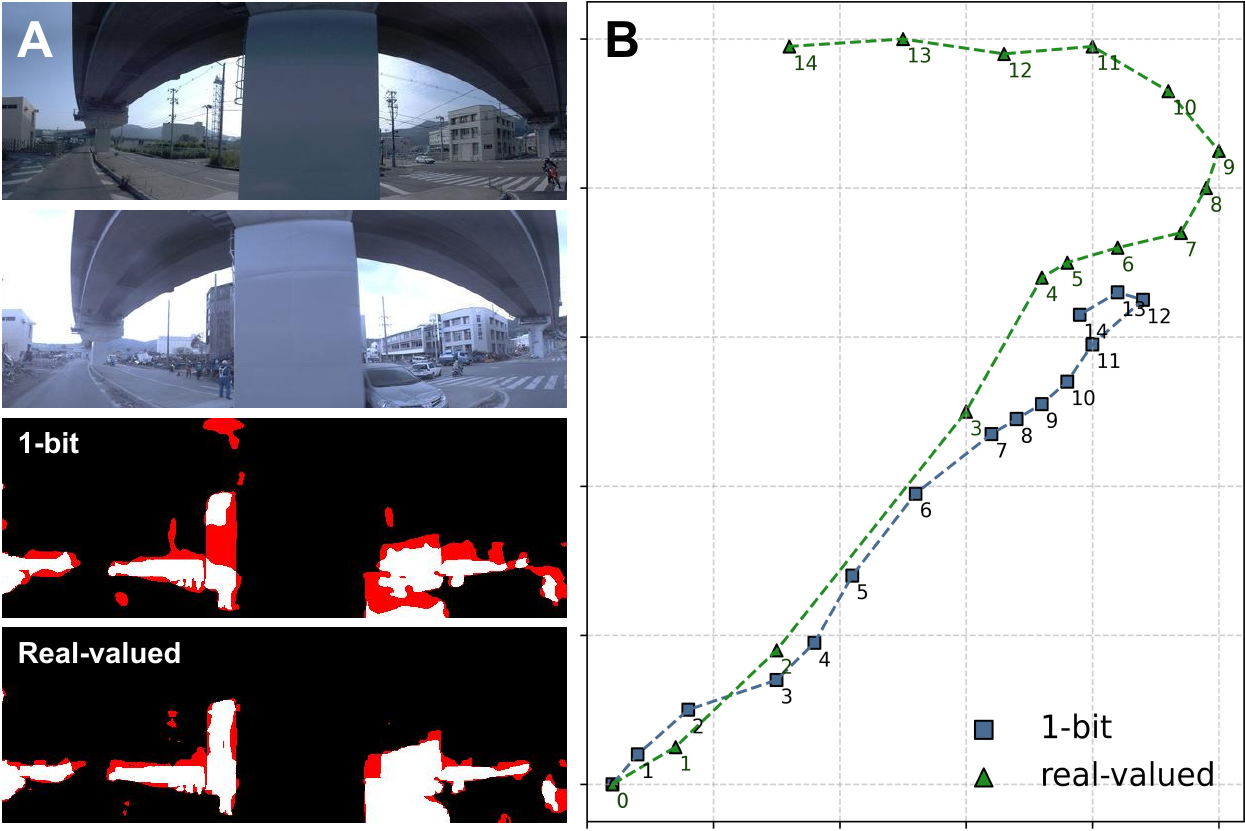}
\vspace{-0.1in}
\caption{
Change detection results on the PCD dataset using a binarized model and real-valued DR-TANet~\cite{Chen2021DRTANetDR}. Left panel (top to bottom): input pair ($t_0$,$t_1$), the error map of the binarized DR-TANet output, and the error map of the real-valued DR-TANet output (white regions: true positives, red regions: true negatives and false positives). The right panel shows the information plane~\cite{shwartz2017opening} dynamics: horizontal axis = $I(X, Z)$ (mutual information between input $X$ and latent feature $Z$), vertical axis = $I(Z, Y)$ (mutual information between latent feature $Z$ and ground truth $Y$). Both networks (15 iterations each) display grey iteration subscripts at the bottom right. Notably, the BNN's $I(X, Z)$ stays not higher than that of the real-valued network at each matched iteration.
}
\label{fig: IB}
\end{figure}

Deep neural networks (DNNs) have significantly advanced change detection performance, enabling applications in urban map updating~\cite{Li2019UrbanBC}, visual surveillance~\cite{Shi2023UnsupervisedCD}, disaster assessment~\cite{Qiao2020ANC}, mobile robotics~\cite{Sofman2011AnytimeON}, and autonomous driving~\cite{He2021DiffNetIF}. However, traditional DNN-based change detection approaches often require substantial computational and storage resources, limiting their deployment on resource-constrained edge devices. To address this, techniques such as model pruning~\cite{xue2024lightweight, zhang2024lightweight}, knowledge distillation~\cite{pang2024hicd, wang2024knowledge}, and compact network design~\cite{feng2023lightweight, wang2024mixcdnet} have been explored in recent years. While these methods enable real-time change detection on edge devices, they still demand non-trivial computational and storage resources.

To further reduce resource usage, network quantization has emerged as a promising solution. By reducing numerical precision to 16/8/4 bits, quantization techniques compress computation and memory demands~\cite{zhuang2020training, Kuzmin2023PruningVQ, Kim2023SqueezeLLMDQ}. At the extreme end lies model binarization, which constrains both weights and activations to 1 bit, achieving up to $32\times$ storage compression and $58\times$ computation reduction~\cite{xnornet}. Binarized networks replace arithmetic operations with energy-efficient logical gates (XNOR and PopCount), making them highly suitable for resource-constrained environments.

Despite its potential, directly applying existing binarization techniques to change detection tasks results in significant performance degradation, as they struggle to distinguish between meaningful ``interest changes", such as the appearance or disappearance of specific objects, and irrelevant ``noise changes" caused by environmental factors. To investigate this issue, we employ mutual information analysis~\cite{tishby2000information, tishby2015deep} to compare binary neural networks (BNNs) and full-precision networks in the information plane~\cite{shwartz2017opening}. As illustrated in Figure~\ref{fig: IB} (B), BNNs exhibit consistently lower mutual information $I(X, Z)$ between input $X$ and hidden representation $Z$ compared to their full-precision counterparts. This limitation stems from the aggressive binarization process, which discards critical feature granularity necessary for robust discrimination in change detection tasks.

\begin{figure*}
\centering
\includegraphics[width=0.85\linewidth]{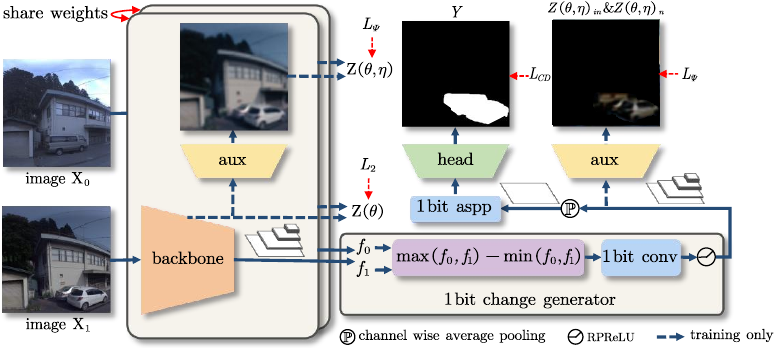}
\caption{BiCD employs a 1-bit C-3PO framework: Dual temporal feature pyramids from a shared 1-bit backbone are merged via a 1-bit change generator. 
Channel-wise average pooling condenses features into a 1-bit ASPP module, outputting a change mask. 
Following Eq.~\ref{eq: bconv}, the 1-bit change generator could be described as $\phi \left( \alpha \odot \left( \left( \max \left( X_0,X_1 \right) -\min \left( X_0,X_1 \right) \right) \circledast W_{}^{b} \right) +\beta \right) $.
Finally, the change detection head outputs a single feature map to predict the change mask.
Features from the 1-bit change generator are fed into the auxiliary module, producing dimension-aligned features $Z(\theta, \eta)_n$ $\&$ $Z(\theta, \eta)_{in}$. 
This process is also applied to dual temporal feature pyramids, ensuring the encoder retains the original input $X$.
}
\label{fig: overview}
\end{figure*}

To address this issue, we propose Binary Change Detection (BiCD), a novel binarized network optimized for change detection using the Information Bottleneck (IB) principle. The IB principle aims to balance feature granularity and discrimination, ensuring that the network retains essential information while discarding irrelevant details. As a fundamental concept in information theory, the IB principle is formally equivalent to the Minimum Description Length (MDL) principle~\cite{Zaidi2020OnTI, Wang2020BiDetAE}, which focuses on data compression. The MDL principle states that the optimal hypothesis is the one that compresses the data most effectively, minimizing two key components: (1) the model description cost and (2) the mismatch between the model and observed data~\cite{Rissanen1978ModelingBS}. This makes the IB principle a powerful tool for model compression and learning compact yet discriminative feature representations~\cite{Louizos2017BayesianCF, Ullrich2017SoftWF, Hu2024ASO, Butakov2023InformationBA}. To achieve this goal, BiCD incorporates an auxiliary objective module guided by the IB principle, enhancing the encoder's ability to retain essential input information while improving feature separability. During training, this module encourages the encoder to focus on distinguishing ``interest changes" from ``noise changes" and preserves more input information through an additional information preservation loss. Importantly, the auxiliary module is removed after training, ensuring no computational overhead during inference. Figure~\ref{fig: overview} shows the proposed BiCD overview.
Our contributions include:
\begin{itemize}
\item[$\bullet$] Binarized Change Detection (\textbf{BiCD}): The first binary neural network specifically designed for change detection. 
Instead of directly applying existing binarization techniques, 
BiCD addresses the unique challenges of change detection by enhancing feature representation and separability.
\item[$\bullet$] \textbf{Auxiliary Objective Module}: A novel module based on the Information Bottleneck principle to overcome the limitations of BNNs in preserving input information. This module guides the encoder to retain essential input information while improving the discrimination between ``interest changes" and ``noise changes," leading to more robust feature extraction.
\item[$\bullet$] \textbf{Efficient and Lightweight Network}: BiCD achieves significant computational and storage efficiency by constraining weights and activations to 1 bit. The auxiliary module is active only during training and removed during inference, ensuring no additional computational overhead.
\item[$\bullet$] Evaluations of the BiCD method across multiple streetscapes and remote sensing change detection datasets. Extensive results demonstrate that our approach achieves an F1-score of 86.5\% on the TSUNAMI subset from the PCD dataset that utilizes a ResNet-18 backbone, thereby establishing a \textbf{new state-of-the-art}.
\end{itemize}
\section{Related Works}
\label{sec:related}
\textbf{Change detection} in paired images is a fundamental problem in computer vision. 
Traditional change detection methods can be broadly categorized into pixel-based and object-based approaches~\cite{Hou2020FromWT}. Pixel-based methods assume precise image registration between paired images and are often sensitive to noise and registration errors. To address these challenges, techniques such as Principal Component Analysis (PCA)~\cite{Deng2008PCAbasedLC} and wavelet transforms~\cite{elik2008UnsupervisedCD} have been widely adopted. In contrast, object-based methods focus on identifying meaningful objects and rely on segmentation algorithms~\cite{Hazel2001ObjectlevelCD, Huo2010FastOC}, offering robustness to registration inaccuracies but requiring more computational resources.
With the advent of deep learning, significant progress has been made in change detection. CDNet~\cite{Sakurada2017DenseOF} leverages bi-temporal images and their estimated optical flow to generate change masks, while CSCDNet~\cite{Sakurada2018WeaklySS} tackles image registration challenges using two correlation layers. DR-TANet~\cite{Chen2021DRTANetDR} introduces a temporal attention module to process bi-temporal image pairs effectively. SimSaC~\cite{Park2022DualTL} employs an optical flow estimation network inspired by GLU-Net~\cite{Truong2019GLUNetGU} to correct distortions between image pairs. More recently, C-3PO~\cite{Wang2022HowTR} develops a network that merges temporal and spatial features to distinguish changes.

\textbf{Auxiliary module} was introduced to address the challenges of training very deep networks~\cite{szegedy2015going}. These modules are active only during training and removed once training is complete, ensuring no additional computational overhead during inference. Recently, auxiliary modules have been widely adopted in local learning frameworks~\cite{wang2024infopro, duan2022training, belilovsky2020decoupled}, where networks are divided into gradient-isolated submodules and trained independently. For instance, Zhang et al.~\cite{zhang2019your} use self-distillation to design multiple network exits, which can be seen as auxiliary modules for Artificial Neural Networks, balancing inference speed and network accuracy. Zhuang et al.~\cite{zhuang2020training} optimize low-bit quantized networks using full-precision auxiliary modules without pre-trained weights. In another study, Xu et al.~\cite{xulearning} enhance a 1-bit tiny remote sensing object detector with an auxiliary decoder. These studies highlight the potential of auxiliary modules to improve network training and performance.

\textbf{Binary neural network} represents the most extreme form of network quantization~\cite{binaryconnect, bnn, QIN2020BNNSurvey}. While this approach achieves significant computational and storage efficiency, binarized models often suffer from reduced accuracy compared to their full-precision counterparts. 
To address this, various techniques have been proposed. For instance, Xnor-Net~\cite{xnornet} introduces scaling factors for weights and activations to mitigate quantization errors. ReActNet~\cite{reactnet} improves BNN performance by incorporating learnable thresholds and RPReLU activation functions to enhance feature representation. Adabin~\cite{Tu2022AdaBinIB} introduces an adaptive binary set to fit diverse data distributions better, increasing the representational flexibility of binarized models. PokeBNN~\cite{zhang2022pokebnn} enhances 1-bit convolutional modules by integrating additional residual connections and a squeeze-and-excitation mechanism to retain critical input information. Despite these advancements, most binarization methods have primarily been evaluated on image classification tasks, leaving their effectiveness in downstream tasks like change detection unexplored~\cite{BinaryDADnet, Wang2020BiDetAE, Cai2023BinarizedSC, Yin2024PathfinderFL}. To bridge this gap, we propose a novel auxiliary module training approach specifically designed for 1-bit change detection networks. This method enhances the network's performance during training without introducing additional inference overhead, offering a tailored solution for the unique challenges of change detection tasks.
\section{Method}
\label{sec:method}
\subsection{Preliminaries}
\textbf{Information Bottleneck (IB)} theory provides a computational framework for balancing data compression and information retention. The mutual information (MI) is a measure of the shared information between two random variables $X$ and $Y$, defined as (for discrete variables)
\begin{equation}
\begin{aligned}
\label{eq: MI}
I\left( X,Y \right) =\underset{x}{\sum{}}\underset{y}{\sum{}}p\left( x,y \right) \log \frac{p\left( x,y \right)}{p\left( x \right) p\left( y \right)} ,
\end{aligned}
\end{equation}
where $p(x,y)$ is the joint probability distribution, and $p(x)$ and $p(y)$ are the marginal distributions. 

The IB theory is a special case of the rate-distortion (RD) theory~\cite{Berger1975, Gibson2025}, which balances lossy compression and distortion. RD theory seeks an optimal compact representation $T$ for a source $X$ via the conditional distribution $p(t|x)$, minimizing the mutual information $I(X, T)$ while ensuring the expected distortion $E(\epsilon)$ remains below a predefined threshold $\epsilon *$. The distortion $d(x, t)$  quantifies the deviation between $X$ and $T$ (e.g., Hamming distance or mean squared error). 
A main limitation of RD theory is the need to predefine the distortion function $d(x,t)$ and the threshold $\epsilon *$, which can be challenging without prior knowledge~\cite{tishby2000information, Hu2024ASO}.  IB theory addresses this by introducing task-relevant ground truth $Y$ and a trade-off parameter $\beta$. Specifically, IB compresses the source $X$ into a compact representation $T$ while maximizing its mutual information with the ground truth of the task, which can be formulated by
\begin{equation}
\begin{aligned}
\label{eq: originIB}
\min \textbf{IB}(p(t|x)) = \min I(X, T) - \beta I(T, Y),
\end{aligned}
\end{equation}
where $I(X, T)$ measures the compression of $X$ into $T$, and $I(T, Y)$ quantifies the retention of task-relevant information. The parameter $\beta$ balances compression and retention, while $p(t|x)$ denotes the mapping from $x$ to $t$.

\textbf{Binary Neural Networks.} In a deep neural network (DNN), let $W$ and $A$ represent the real-valued weights and activations. For the $i$-th layer, which typically ends with a nonlinear mapping $\phi$ (e.g., ReLU) in an inner product space, the convolution is expressed as $A_{i+1}=\phi \left( A_i\otimes W_i \right)$, where $\otimes$ is the standard convolution operation.

\begin{figure}
\centering
\includegraphics[width=3.2in, keepaspectratio]{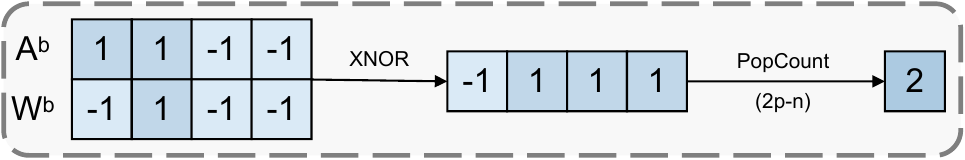}\\
\caption{ PopCount counts the number of ``1"s in a binary-state vector (-1s or 1s). After the XNOR operation, $p$ represents the number of set bits (i.e., the PopCount result), and $u$ represents the number of unset bits (i.e., the count of ``-1"). $n$ denotes the vector length ($n=p+u$). By using $u=n-p$, we can express the dot product result as $p-(n-p)$.}
\label{fig: xnorpopcount}
\end{figure}

BNNs use the sign function to constrain both weights and activations to \{-1, +1\}~\cite{xnornet}, and use the Straight-Through Estimator(STE)~\cite{Bengio2013STE} to alleviate the sign function's non-differentiability. To better approximate real values $x \in \mathbb{R}^n$, BNNs introduce learnable scaling factors $\alpha_w$ and $\alpha_a$ for weights and activations, respectively~\cite{xnornet}, and an additional learnable bias $b$~\cite{reactnet} to reduce quantization errors. For simplicity, we consolidate the scaling factors for weights and activations into a single parameter $\alpha$~\cite{Bulat2019XNORNetIB}. 
Thus, the binary convolution operation can be formulated as:
\begin{equation}
\begin{aligned}
\label{eq: bconv}
A_{i+1}\approx \phi \left( \alpha \odot \left( A_{i}^{b}\circledast W_{i}^{b} \right) +\beta \right),
\end{aligned}
\end{equation}
where $\circledast$ denotes a bit-wise operation that includes the XNOR-PopCount operation, which replaces the floating-point matrix multiplication in traditional convolution operations, significantly accelerating computation, as shown in Figure~\ref{fig: xnorpopcount}. Here, $\odot$ denotes element-wise multiplication.

\subsection{Information Bottleneck for Change Detection}
Tishby et al.~\cite{tishby2015deep} proposed that the hierarchical structure of DNNs can be viewed as a continuous Markov chain, representing all relationships and data~\cite{shwartz2017opening}.
For change detection tasks, the intermediate layers of a CNN network extract latent features $Z=f\left( X,\theta \right) $, where $X$ represents the input image pairs. These features are discriminative across ``interest change" and ``noise change." The learned latent features complete the change detection process through the decoder $g\left( Z \right)$ to predict the label Y:
\begin{equation}
\begin{aligned}
\label{eq: markov}
X\xrightarrow{f\left( X,\theta \right)}Z\left( \theta \right) \xrightarrow{g\left( Z \right)}Y.
\end{aligned}
\end{equation}

The Information Bottleneck (IB) principle hypothesizes that DNNs learn latent features $Z$ as a minimal sufficient statistic for label $Y$. Following Eq.~\ref{eq: originIB}, IB seeks to maximize the mutual information $I (Z, Y)$ between $Z$ and $Y$ while minimizing the mutual information $I (X, Z)$ between $X$ and $Z$~\cite{shwartz2017opening}:
\begin{equation}
\resizebox{0.90\linewidth}{!}{%
    $\begin{aligned}
    \label{eq: IB}
    \min \textbf{IB}(X, Y, Z(\theta)) = \min I(X, Z(\theta)) - \beta I(Z(\theta), Y),
    \end{aligned}$%
}
\end{equation}
where $I(\cdot, \cdot)$ denotes the mutual information between two random variables. $I\left(X, Z(\theta)\right)$ represents the compression of X, limiting the amount of information the decoder can extract. $I\left(Z(\theta), Y\right)$ represents the similarity with Y, enabling the decoder to retain more information relevant to the task. However, as shown in Figure~\ref{fig: IB}, binary neural networks (BNNs) exhibit significantly lower $I(X, Z(\theta))$ compared to full-precision networks due to their limited feature extraction capability. This reduction in $I(X, Z(\theta))$ essentially impacts the ``drift phase" ~\cite{shwartz2017opening}of network training, which, in turn, leads to a lower $I(Z(\theta), Y)$. For change detection tasks, this constraint hinders their ability to distinguish between  ``interest change" and ``noise change", making it challenging to achieve optimal performance during training. Specifically, maximizing $I(Z(\theta), Y)$ becomes difficult, as the network cannot retain sufficient discriminative information.

To address this limitation, unlike previous works~\cite{yin2024si, yin2024bidense, xin2025biisnetcvpr} that focus solely on improving the feature extraction capability of BNNs from input $X$, we introduce an auxiliary objective to improve feature separability in hidden representations. This auxiliary objective enables the encoder to better distinguish between ``interest changes" and ``noise changes" in change detection tasks, simplifying the decoder's job. We quantify separability through spatial distance in hidden features. For encoder $f$ with nonlinear mapping $\phi$ in an inner product space: $f\left( \cdot \,\,,\,\,\theta \right) =\phi \left( g\left( \cdot ,\theta \right) \right),$ where $g$ is any function can help learn parameters $\theta$, the separability is defined as following: 
\begin{equation}
\begin{aligned}
\label{eq: separability}
\left\{ \begin{array}{l}
	\max\text{\,\,}||\phi \left( g\left( x_i,\theta \right) \right) -\phi \left( g\left( x_j,\theta \right) \right) \,\,||,\\
    \ \ \ \ \ \ \ \ \ \ \ \ \ \ \ \ \  \ \ \  \ \    if\,\,\,\,x_i,\,\,x_j\,\,from\,\,X\,\,with\,\,y_i\ne y_j\\
	\min\text{\,\,}||\phi \left( g\left( x_i,\theta \right) \right) -\phi \left( g\left( x_j,\theta \right) \right) \,\,||,\\
    \ \ \ \ \ \ \ \ \ \ \ \ \ \ \ \ \ \ \  \ \ \  if\,\,\,\,x_i,\,\,x_j\,\,from\,\,X\,\,with\,\,y_i=y_j\\
\end{array} \right. 
\end{aligned}
\end{equation}

Following the IB principle, we have redesigned our computational framework as defined in the Eq.~\ref{eq: separability}:
\begin{equation}
\begin{aligned}
\label{eq: newIB}
&\min\, I(X, Z(\theta)) - \beta I(Z(\theta), Y, \Delta X)=\\
\min\, &I(X, Z(\theta)) - \beta_1 I(Z(\theta), Y) - \beta_2 I(Z(\theta), \Delta X),
\end{aligned}
\end{equation}
where $\Delta X$ represents the absolute difference between input images $X_0$ and $X_1$:
$\Delta X = |X_0 - X_1|.$
This computation applies element-wise absolute value operations to corresponding pixel pairs.

Due to the network-imposed limitations on the dimensions of latent features $Z$, estimating mutual information becomes challenging when the dimensions of $Z$ do not match those of the input $X$ and the label $Y$. 

\subsection{Auxiliary Objective Module}
To enable flexible adjustment of hidden feature dimensions for auxiliary goal computation, we employ an Auxiliary Module~\cite{pyeon2020sedona, wang2024infopro}, a technique widely used in local learning. This module maps features between different feature spaces through dimension conversion while controlling error margins~\cite{wangrevisiting, duan2022training, xulearning}. Given the inherent lightweight inference and training sensitivity of binary neural networks (BNNs), the compact auxiliary network's training overhead is justified, as it incurs no inference cost (the module is removed during deployment).
We define the auxiliary module as $\sigma (\cdot, \eta)$, where $\eta$ denotes its trainable parameters. The auxiliary module consists of four parallel MLP branches, followed by a convolutional output layer to generate compressed representations.
This allows us to reformulate the separability objective as follows:

\begin{equation}
\begin{aligned}
\label{eq: separabilitynew}
\left\{ \begin{array}{l}
	\max\text{\,\,}||\sigma \left( \phi \left( g\left( x_i,\theta \right) \right), \eta \right) - \sigma \left( \phi \left( g\left( x_j,\theta \right) \right), \eta \right) \,\,||,\\
    \ \ \ \ \ \ \ \ \ \ \ \ \ \ \ \ \ \ \ \ \  \ \ \ \ \ \ \ \    if\,\,\,\,x_i,\,\,x_j\,\,from\,\,X\,\,with\,\,y_i\ne y_j\\
	\min\text{\,\,}||\sigma \left( \phi \left( g\left( x_i,\theta \right) \right), \eta \right) - \sigma \left( \phi \left( g\left( x_j,\theta \right) \right), \eta \right) \,\,||,\\
    \ \ \ \ \ \ \ \ \ \ \ \ \ \ \ \ \ \ \ \ \ \ \ \ \ \ \ \ \  if\,\,\,\,x_i,\,\,x_j\,\,from\,\,X\,\,with\,\,y_i=y_j\\
\end{array} \right. 
\end{aligned}
\end{equation}
where $\theta$ and $\eta$ are jointly optimized during training. The auxiliary module $\sigma$ is discarded during inference, adding no extra computational cost. 
With the addition of an auxiliary module, our computational framework expands to:
\begin{equation}
\begin{aligned}
\label{eq: latestIB}
\min\,\,I\left( X,Z\left( \theta \right) \right) -\beta _1I\left( Z\left( \theta \right) ,Y \right) - \beta_2 \varPsi,
\end{aligned}
\end{equation}
where $\varPsi$ formulates the auxiliary objective:
\begin{equation}
\begin{aligned}
\label{eq:auxmoduleIB1}
\varPsi = I(Z(\theta, \eta), \Delta X) + I(X, Z(\theta, \eta)),
\end{aligned}
\end{equation}
where $Z(\theta, \eta)$ represents the aligned latent feature.
In our framework, $I(Z(\theta, \eta), \Delta X)$ represents the feature separability improvement from our auxiliary objective. This mutual information metric directly measures distinguishable feature learning capacity.
The term $I(X, Z(\theta, \eta))$ represents the reconstruction loss for dimension alignment of the auxiliary module, which encourages the encoder to retain input $X$. This conceptually contrasts with the IB principle that minimizes $I(X, Z(\theta))$. 
Notably, $I(Z(\theta, \eta), \Delta X)$ serves dual purposes. First, it acts as the reconstruction loss for the auxiliary module in change feature dimension alignment. Second, it functions as the reconstruction loss for the auxiliary module in the Siamese network for original feature extraction.

\subsection{Mutual Information Estimation}
We detail our approach for estimating mutual information $I(X, Z(\theta))$, $I(Z(\theta), Y)$ (external to $\varPsi$), and $I(Z(\theta, \eta), \Delta X)$ within $\varPsi$. These estimations culminate in the final learning objective for our network.
Following the definition of mutual information in Eq.~\ref{eq: MI}, we reformulate $I(X, Z(\theta))$
using the Kullback-Leibler divergence:
\begin{equation}
\begin{aligned}
\label{eq: IXZ}
&I\left( X,Z(\theta) \right) =D_{KL}\left[ p\left( x,z \right) ||p\left( z \right) \right] \\
=&\underset{x\in X}{\sum{}}p\left( x \right) \underset{z\in Z(\theta)\,\,}{\sum{}}p\left( z|x \right) \,\,\log \left( \frac{p\left( z|x \right)}{p\left( z \right)} \right) 
\end{aligned}
\end{equation}
Building upon the engineering approximation framework established in prior work~\cite{Wang2020BiDetAE, bidetpage}, we adopt L2-norm regularization to minimize $I(X, Z(\theta))$, avoiding direct computation of the intractable posterior $p(z|x)$.

For the external mutual information term $I(Z(\theta), Y)$, this term corresponds to the original change detection loss function, as defined in~\cite{Sakurada2018WeaklySS}.
For the internal term $I(Z(\theta, \eta), \Delta X)$, we first generate change features using a 1-bit convolution operation, referred to as the ``1-bit change generator" (see Figure~\ref{fig: overview}). These features are used to produce aligned features $Z(\theta, \eta)$. Then we apply element-wise multiplication between change mask $Y$ and aligned features $Z(\theta, \eta)$, yielding ``noise change" component $Z(\theta, \eta)_{\text{n}}$ and ``interest change" component $Z(\theta, \eta)_{\text{in}}$.
$\Delta X_{\text{in}}$ is generated analogously without ``1-bit generator" and align step. To enhance feature separability, we reformulate $\varPsi$ as:
\begin{equation}
\begin{aligned}
\label{eq:auxmoduleIB2}
\varPsi  = I(Z(&\theta, \eta)_{\text{n}}, 0) + I(Z(\theta, \eta)_{\text{in}}, \Delta X_{\text{in}}) \\
&+ I(X, Z(\theta, \eta)),
\end{aligned}
\end{equation}
where $I(Z(\theta, \eta)_{\text{n}}, 0)$ suppresses ``noise changes" via denoising-like operations, while $I(Z(\theta, \eta)_{\text{in}}, \Delta X_{\text{in}})$ preserves ``interest changes" during denoising. Taking $I(X, Z(\theta, \eta))$ as an example, let $R(X|Z(\theta, \eta))$ denote the expected reconstruction error from $Z$ to $X$. Following mutual information definitions, we implicitly formulate it as:
\begin{equation}
\begin{aligned}
I\left( X, Z(\theta,\eta) \right) &= H\left( X \right) - H\left( X | Z(\theta,\eta) \right) \\
&\ge H\left( X \right) - R\left( X | Z(\theta, \eta) \right)
\end{aligned}
\label{eq:recons_error}
\end{equation}
where $H(\cdot)$ denotes entropy, and $H(X)$ is a constant~\cite{hjelmlearning}. The auxiliary module $\sigma$ parameterized by $\eta$ estimates $I(X, Z(\theta,\eta)$ to minimize reconstruction loss: $I(X,Z(\theta,\eta))\approx \max \left[ H(X) - R_{\sigma}(X|Z(\theta,\eta)) \right]$. Consequently, the three mutual information terms in $\varPsi$ represent reconstruction losses for auxiliary sub-networks. Following~\cite{zhao2016loss}, we adopt L1 loss in implementation.

To clarify the formula's meaning, we repositioned the hyperparameters in the equation. The final objective of our BiCD is defined as:
\begin{equation}
\begin{aligned}
\label{eq: finalObj}
\min &\text{ Obj} = \beta_1 \|Z(\theta)\|_2 +  L_{\text{cd}} + \beta_2 ( \|Z(\theta, \eta)_n\|_1 \\
&+ \|Z(\theta, \eta) - X\|_1 +  \|Z(\theta, \eta)_{\text{in}} - \Delta X_{\text{in}}\|_1 ),
\end{aligned}
\end{equation}
where $\|Z(\theta)\|_2$ corresponds to $I(X, Z(\theta))$, $L{\text{cd}}$ represents $I(Z(\theta), Y)$ (the change detection task loss), and the remaining terms derive from $\varPsi$. Hyperparameters $\beta_1$ and $\beta_2$ balance these terms, ensuring optimal performance.
\section{Experiment}
\label{sec:experiment}
In this section, we evaluate the effectiveness of our BiCD method on three widely used change detection datasets: PCD, VL\_CMU\_CD, and LEVIR-CD. These datasets cover street-view and remote sensing scenarios, ensuring a comprehensive assessment of our method's generalization capability. We conduct extensive experiments to validate the performance of BiCD, including comparisons with state-of-the-art binary and full-precision networks, ablation studies to analyze the impact of key components and hyperparameters, and latency measurements on edge devices.

\textbf{Training Details.} We use the Adam~\cite{kingma2014adam} optimizer with a starting learning rate of 5e-4 and cosine annealing. Training runs on a single RTX 3090 GPU (batch size=4). Following the previous work~\cite{Wang2022HowTR}, the backbone is pre-trained on ImageNet-1K~\cite{imagenet}, while other modules start from random values. The initial learning rate for auxiliary modules is 5e-3, with the same warmup as the backbone. Their learning rate drops to 1/10 at epochs 90 and 120. All models train for 140 epochs with fixed settings.

\textbf{PCD dataset}~\cite{sakurada2015change} is designed for street scene change detection. This dataset includes two subsets, TSUNAMI and GSV, each containing 100 panoramic image pairs (original size 224×1024) with manually labeled change masks as ground truth. Following prior work \cite{Lei2020HierarchicalPC, Chen2021DRTANetDR, Wang2022HowTR}, we crop images using a 56-pixel sliding window to generate 3,000 paired tokens of 224×224 resolution. These tokens are augmented by rotating them at 0°, 90°, 180°, and 270°, resulting in 12,000 total samples. We implement 5-fold cross-validation, with each training set containing 9,600 augmented tokens and each test set using 20 original images.

\textbf{VL\_CMU\_CD dataset}~\cite{Alcantarilla2016StreetviewCD} contains 152 perspective image sequences for change detection, with 1,362 image pairs and manual ground truth annotations. The original 1024x768 images are resized to 512x512. Following prior work~\cite{Varghese2018ChangeNetAD, Wang2022HowTR}, we split the dataset into 933 training pairs and 429 test pairs. Through rotation augmentation, the training set expands to 3,732 pairs. We train models on this augmented set and report metrics on the test set.

\textbf{LEVIR-CD dataset}~\cite{Chen2020ASA} is designed for remote sensing scene change detection. It contains 637 image pairs (1024$\times$1024 pixels, 0.5m resolution) collected via Google Earth API, focusing on building change detection. Following~\cite{Chen2021RemoteSI, Li2022RemoteSC, Ying2024DGMA2NetAD}, we randomly split the data into training/validation/test sets (7:1:2 ratio). All images are cropped to 256$\times$256 pixels, producing 7,120 training pairs, 1,024 validation pairs, and 2,048 test pairs. 

\begin{table}[t]
\centering
\caption{Quantitative analysis of real-valued change detection models, 1-bit change detection models, and our proposed methods. The evaluation covers the F1-score on the PCD dataset.}
\label{tab:pcd}

\renewcommand{\arraystretch}{1.0}
\setlength{\tabcolsep}{0.5pt}
\scalebox{0.9}{
\begin{tabular}{cccccp{2cm}}  
\toprule
Framework & Method & \small Bits & \multicolumn{1}{c}{\begin{tabular}[c]{@{}c@{}}\small Params.\\ \small (M)\end{tabular}} & \multicolumn{1}{c}{\begin{tabular}[c]{@{}c@{}}\small OPs\\ \small (G)\end{tabular}} & \multicolumn{1}{c}{\begin{tabular}[c]{@{}c@{}}F1-score(\%)\\ \scriptsize GSV \ TSUNAMI \end{tabular}} \\
\midrule

FC-Siam-di~\cite{Chen2021RemoteSI}
  & \small Real-valued      
  & 32     
  & 1.4                
  & 4.7                
  & \ 66.2 \  \ 79.5 \\

\midrule
FC-Siam-co~\cite{Chen2021RemoteSI}
  & \small Real-valued      
  & 32     
  & 1.6               
  & 5.3
  & \ 70.4 \  \ 81.6 \\ 

\midrule
CSCDNet~\cite{Sakurada2018WeaklySS}
  & \small Real-valued      
  & 32     
  & 42.2               
  & 94.2
  & \ 72.8 \  \ 87.8\\

\midrule
\multirow{5}*{DR-TANet~\cite{Chen2021DRTANetDR}}
  & \small Real-valued      
  & 32 
  & 33.4      
  & 28.5
  & \ 72.3 \ \ 87.6\\
  \cdashline{2-6}

  & \small BNN          
  & 1
  & \multirow{4}{*}{1.1}
  & \multirow{4}{*}{3.0}
  & \ 50.1  \ \ 69.3\\

  & \small ReActNet         
  & 1
  & 
  & 
  & \ 65.7  \ \ 83.4\\
  
  & \small AdaBin          
  & 1
  & 
  & 
  & \ 64.3 \ \ 82.0\\

  & \small BiCD (Ours)           
  & 1
  & 
  & 
  & \ 67.7  \ \ \textbf{85.1}\\

\midrule 
\multirow{5}*{C-3PO~\cite{Wang2022HowTR}}
  & \small Real-valued      
  & 32 
  & 40.5      
  & 229.0
  & \ 77.8  \ \ 88.4 \\
  \cdashline{2-6}

  & \small BNN           
  & 1
  & \multirow{4}{*}{2.1}
  & \multirow{4}{*}{6.6}
  & \ 50.9  \ \ 71.7\\

  & \small ReActNet           
  & 1
  & 
  & 
  & \ 71.2  \ \ 84.0\\
  
  & \small AdaBin           
  & 1
  &
  &
  & \ 70.1 \ \ 82.8\\

  & \small BiCD (Ours)           
  & 1
  &
  &
  & \ 74.1  \ \ \textbf{86.5}\\

\bottomrule
\end{tabular}}
\end{table}
\subsection{Comparison with SOTA BNNs}
As the first work to explore BNNs for change detection, we compare our BiCD with existing BNNs (e.g., BNN~\cite{bnn}, ReAct-Net~\cite{reactnet}, AdaBin~\cite{Tu2022AdaBinIB}) which have achieved state-of-the-art (SOTA) performance on the classification task. We rerun the code of these SOTA BNNs supplied by authors on the change detection task and report the F1-score of the best epoch. We choose the latest change detection methods, DR-TANET~\cite{Chen2021DRTANetDR} and C-3PO\cite{Wang2022HowTR}, as the frameworks and integrate our BiCD and exiting SOTA BNNs to these frameworks for evaluation.
The results of full-precision networks are also included as references. 
Memory usage follows the mainstream approach~\cite{reactnet, Wang2020BiDetAE}: 32$\times $ full-precision kernels + 1$\times $ binary kernels. Operations (OPs) are calculated as real-valued FLOPs plus 1/64 of 1-bit multiplications, following Bi-Real-Net's protocol~\cite{birealnet}.  
BiCD leverages CPU parallel bit-wise operations (XNOR and PopCount) for significant acceleration and memory reduction while maintaining competitive accuracy.

\textbf{Results on PCD.}
Table~\ref{tab:pcd} compares computational complexity, memory costs, and F1-score across quantization methods and change detection frameworks. 
BiCD delivers remarkable efficiency gains: 9.5× fewer operations with 30.3× memory savings in DR-TANet and 34.7× operation reduction alongside 19.3× memory compression in 1-bit C-3PO. The 1-bit C-3PO's efficiency breakthrough addresses its real-valued counterpart's computational bottleneck. The original reduction convolution layer (processing 128×128 multi-scale spatial features) required 0.155T operations. We replace these with parameter-free channel-wise average pooling, eliminating computational overhead while preserving spatial feature integration capabilities.

When integrated with DR-TANet and C-3PO, BiCD outperforms SOTA BNNs by 1.7\% / 2.5\% F1-score on the TSUNAMI subset and 2.0\% / 2.9\% on the GSV subset, without additional computational/memory costs. Remarkably, BiCD with 1-bit C-3PO surpasses full-precision DR-TANet while matching the real-valued DR-TANet / C-3PO performance on the TSUNAMI subset. These results demonstrate critical implications for real-time change detection networks.

\textbf{Results on VL\_CMU\_CD.}
The VL\_CMU\_CD dataset poses more significant challenges for change detection than the PCD dataset due to its diverse change types and scales. As shown in Table~\ref{tab:vlcmucd}, our BiCD method outperforms state-of-the-art binary neural networks, achieving F1-score improvements of 3.3\% and 2.0\% in DR-TANet and C-3PO frameworks, respectively, through enhanced separability of hidden features and implicit information retention. Under the 1-bit C-3PO framework, BiCD attains performance comparable to the real-valued CSCDNet model.

\begin{table}[t]  
\centering
\caption{Quantitative analysis of real-valued change detection models, 1-bit change detection models, and our proposed methods. The evaluation covers the F1-score on the VL\_CMU\_CD dataset.}
\label{tab:vlcmucd}

\renewcommand{\arraystretch}{1.0}
\setlength{\tabcolsep}{3pt}
 \scalebox{0.9}{
\begin{tabular}{cccccc}  
\toprule
Framework & Method & Bits  & \multicolumn{1}{c}{\begin{tabular}[c]{@{}c@{}}F1-score (\%)\end{tabular}} \\

\midrule
FC-Siam-di~\cite{Chen2021RemoteSI}
  & Real-valued      
  & 32     
  & 65.3                \\

\midrule
FC-Siam-co~\cite{Chen2021RemoteSI}
  & Real-valued      
  & 32     
  & 65.6                \\
 
\midrule
CSCDNet~\cite{sakurada2015change}
  & Real-valued      
  & 32     
  & 71.0                \\

\midrule
\multirow{5}*{DR-TANet~\cite{Chen2021DRTANetDR}}
  & Real-valued      
  & 32       
  & 73.2                \\

\cdashline{2-6} 
  & BNN           
  & 1
  & 44.4             \\
  
  & ReActNet           
  & 1
  & 62.6               \\
  
  & AdaBin           
  & 1
  & 59.6               \\

  & BiCD (Ours)           
  & 1
  & \textbf{65.9  }             \\

\midrule 
\multirow{5}*{C-3PO~\cite{Wang2022HowTR}}
  & Real-valued      
  & 32    
  & 79.5                \\

\cdashline{2-6} 
  & BNN           
  & 1
  & 49.3               \\
  
  & ReActNet           
  & 1
  & 69.9               \\
  
  & AdaBin           
  & 1
  & 67.0               \\

  & BiCD (Ours)           
  & 1
  & \textbf{71.9}               \\

\bottomrule
\end{tabular}}
\end{table}

\begin{table}[t]
\centering
\caption{Quantitative analysis of real-valued remote sensing models, real-valued lightweight remote sensing models, real-valued street-view models, and our BiCD. The evaluation includes the number of parameters, Operation Quantity (OPs), and F1-score on the LEVIR-CD dataset.}
\label{tab:levir-cd}

\setlength{\tabcolsep}{5.5pt}%
\renewcommand{\arraystretch}{1.0}%
 \scalebox{0.9}{
\begin{tabular}{lcccc}
\toprule
Method & Bits
 & \multicolumn{1}{c}{\begin{tabular}[c]{@{}c@{}}Params.\\(M)\end{tabular}}
 & \multicolumn{1}{c}{\begin{tabular}[c]{@{}c@{}}OPs\\(G)\end{tabular}}
 & \multicolumn{1}{c}{\begin{tabular}[c]{@{}c@{}}F1-score\\(\%)\end{tabular}} \\
\midrule
EGRCNN~\cite{bai2021edge}       & 32 & 9.6   & 17.6 & 89.6 \\

DARNet~\cite{Li2022ADA}       & 32 & 15.1  & 64.2 & 90.6 \\

EGCTNet~\cite{Xia2022BuildingCD}    & 32 & 106.1  & 33.5 & 90.7 \\

FC-Siam-di~\cite{Daudt2018FullyCS} & 32 & 1.4   & 4.7  & 84.4 \\

FC-Siam-co~\cite{Daudt2018FullyCS} & 32 & 1.6   & 5.3  & 81.8 \\

BIT~\cite{Chen2021RemoteSI}          & 32 & 12.4   & 10.9 & 89.3 \\

Changeformer~\cite{Bandara2022ATS}          & 32 & 41.0   & 46.0 & 90.4 \\

DDPM-CD~\cite{Bandara2025DDPMCDDD}          & 32 & 46.4   & 2182.0 & 90.9 \\

M-CD~\cite{Paranjape2024AMS}          & 32 & 69.8   & 29.6 & 92.1 \\

MaskCD~\cite{Wang2022HowTR}        & 32 & 107.4  & -- & 90.9 \\

C-3PO~\cite{Wang2022HowTR}        & 32 & 40.5  & 222.0& 90.9 \\
\midrule
\begin{tabular}[l]{@{}l@{}}
    1bit C-3PO\\
    +ReActNet
\end{tabular}    & 1  &   2.1    &   6.6   & 88.8 \\
\midrule
\begin{tabular}[l]{@{}l@{}}
    1bit C-3PO\\
    +BiCD (Ours)
\end{tabular} 
& 1 &  2.1    &  6.6    & \textbf{89.9} \\
\bottomrule
\end{tabular}}
\vspace{-0.05in}
\end{table}
\textbf{Results on LEVIR-CD.}
Remote sensing change detection fundamentally differs from street-view tasks in data characteristics~\cite{wang2023changes}. Precisely aligned imagery eliminates viewpoint variations while exhibiting simpler noise patterns. However, it poses unique challenges. As shown in table~\ref{tab:levir-cd}, our method demonstrates 1.1\% F1-score improvement over state-of-the-art BNNs. Compared to a real-valued counterpart, BiCD achieves comparable performance while outperforming some lightweight full-precision counterparts. These findings validate the necessity of 1-bit networks for change detection in remote sensing applications.

\begin{figure}[t]
\centering
\includegraphics[width=3.2in, keepaspectratio]{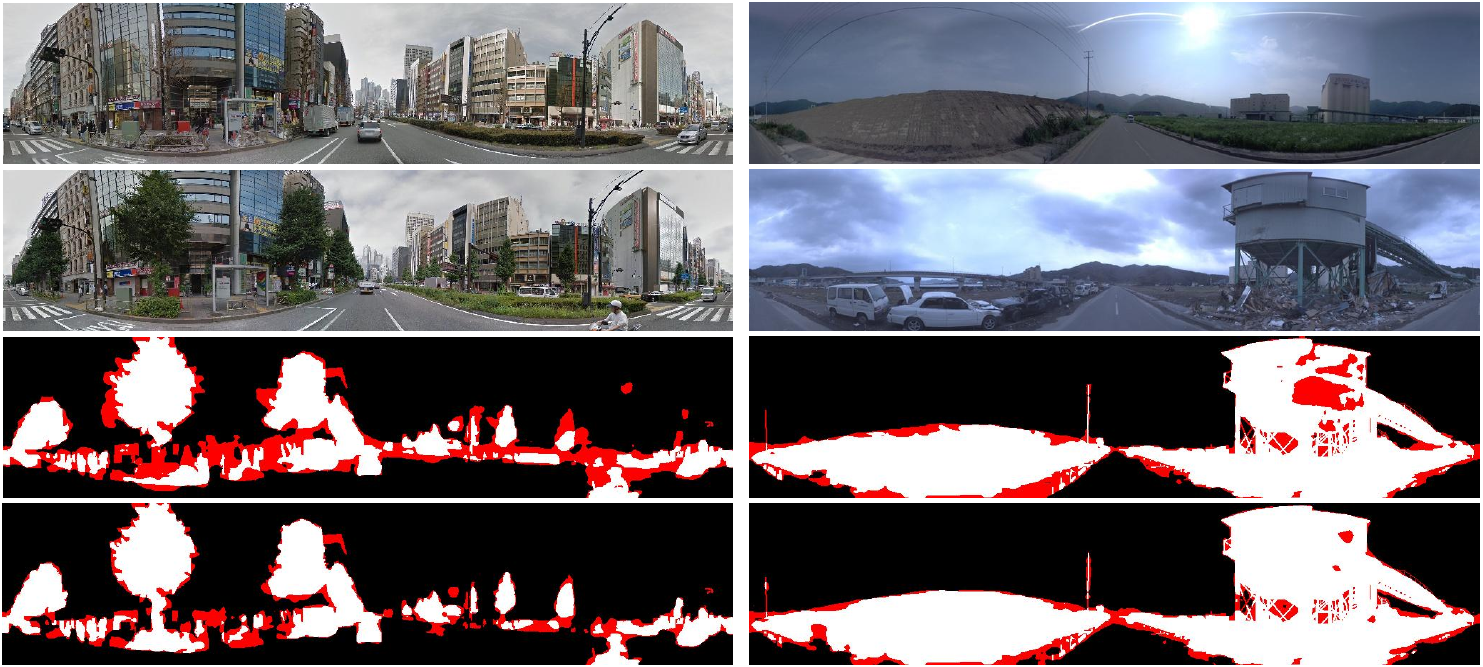}\\
\caption{Change detection results on GSV (left) and TSUNAMI (right) subsets of the PCD dataset, comparing the 1-bit baseline and BiCD based on the C-3PO framework. From top to bottom: input pair ($t_0$, $t_1$), error map of the 1-bit baseline output, and error map of the BiCD output.}
\label{fig: vis_pcd}
\end{figure}
\begin{figure}[t]
\centering
\includegraphics[width=2.7in, keepaspectratio]{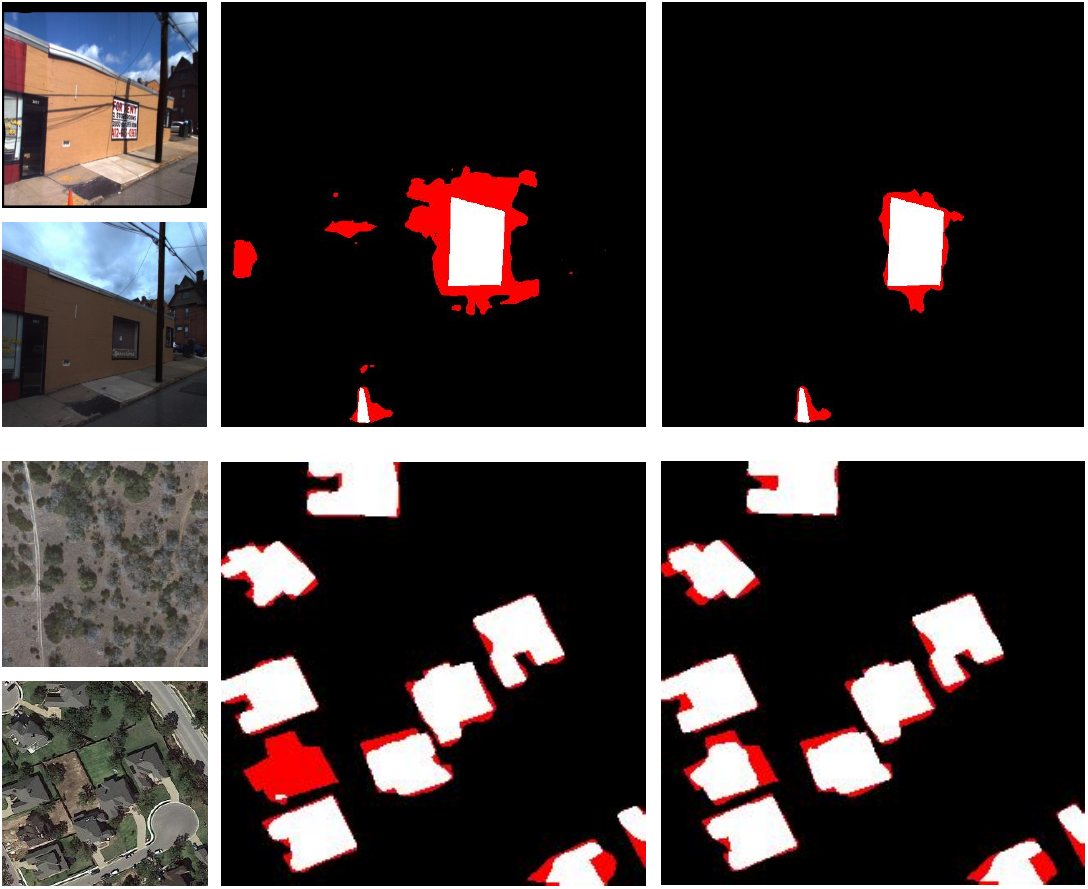}\\
\caption{Change detection results on VL\_CMU\_CD (top) and LEVIR-CD (bottom) datasets, comparing 1-bit baseline and BiCD based on the C-3PO framework. Left to right: input pair ($t_0$ \& $t_1$), error map of 1-bit baseline output, and error map of BiCD output.}
\label{fig: vis_cmu_rs}
\end{figure}
\textbf{Visualization.}
Qualitative results on the above three datasets are shown in Figure~\ref{fig: vis_pcd} and Figure~\ref{fig: vis_cmu_rs}, from which we can clearly see that our BiCD significantly outperforms existing 1-bit baselines, accurately detecting the changes with much fewer errors (white regions: true positives; red regions: true negatives and false positives). This improvement mainly stems from our BiCD's enhanced feature separability, enabling us to effectively distinguish between meaningful ``interest changes" and irrelevant ``noise changes."

\subsection{Ablation Studies}

\begin{figure}
\centering
\includegraphics[width=3.2in, keepaspectratio]{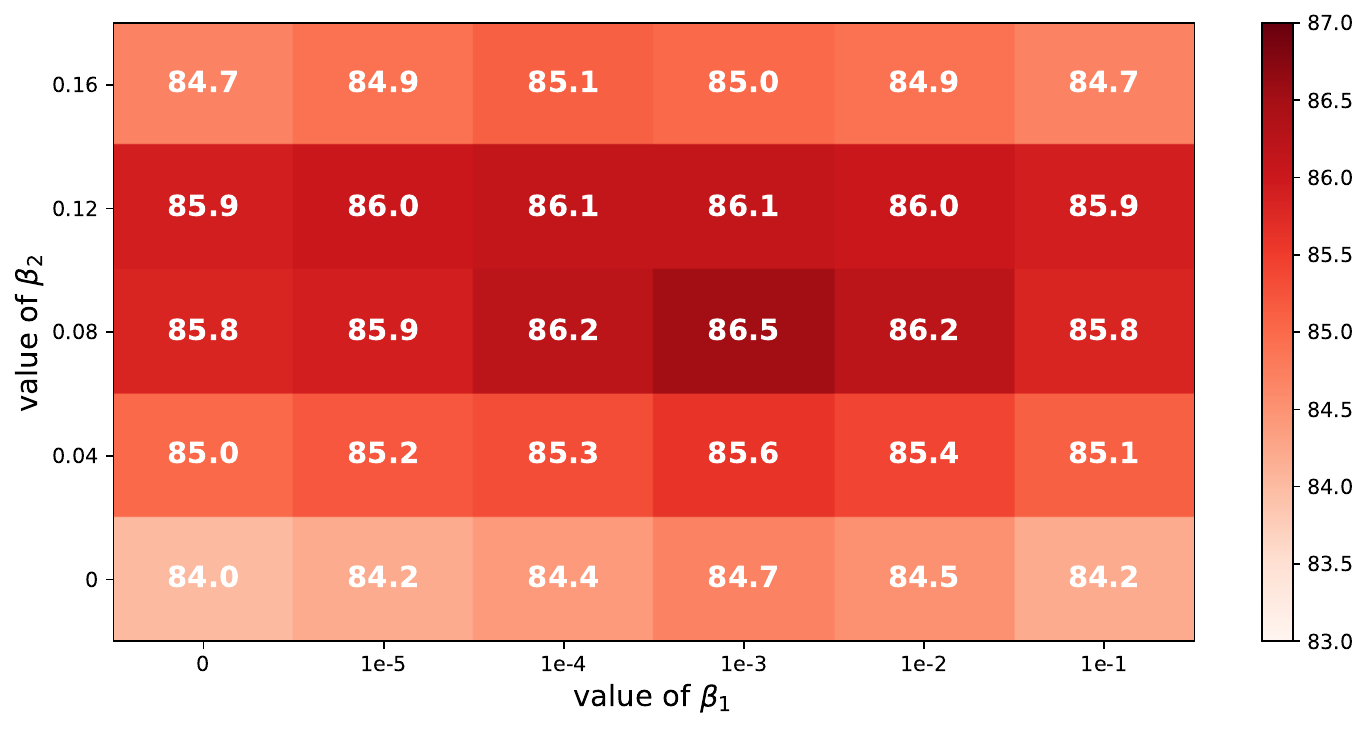}\\
\caption{The hyper-parameters $\beta_1$ and $\beta_2$ in our BiCD framework are optimized through empirical analysis of the PCD dataset's TSUNAMI subset.}
\label{fig: parameter}
\end{figure}

\textbf{Hyper-Parameter Selection.}
We select hyperparameters $\beta_1$ and $\beta_2$ using the C-3PO framework with the ResNet-18 backbone. Figure~\ref{fig: parameter} presents the F1-score performance under different $\beta_1$ and $\beta_2$ settings. 
Since $\beta_1$ controls the suppression rate of task-irrelevant redundant information in binarized change detection networks, the 1-bit C-3PO achieves optimal performance at $\beta_1=1e-3$. We, therefore, fix $\beta_1=1e-3$ for extended ablation studies. For $\beta_2$ governing feature separability, the original baseline ($\beta_2=0$) underperforms all BiCD variants ($\beta_2>0$), demonstrating the necessity of our method. Experimental results show that $\beta_2=0.08$ delivers the best performance, achieving 86.5\% F1-score on the TSUNAMI subset of the PCD dataset with C-3PO. Based on these findings, we set $\beta_2=0.08$.

\textbf{Effective of Components.}
Table~\ref{tab: ablation} summarizes the component-wise improvements in BiCD, with gains reported relative to the baseline F1-score of 84.7\%. Key findings include:
(1) In the Siamese network, adding the full auxiliary objective (including separability) yields only +0.1\%, while removing the separability term $I(Z(\theta,\eta), \Delta X)$ improves performance by +0.5\%. This suggests the reconstruction loss $I(X, Z(\theta, \eta))$ implicitly preserves input information. However, directly imposing the separability objective inside the Siamese network degrades performance.
(2) Transferring the complete auxiliary objective, including the separability objective, to the 1-bit generator achieves a 1.1\% improvement, but removing separability eliminates gains. This outcome suggests that successful separability optimization requires direct interaction with change-specific features in the generator. Meanwhile, implicit preservation of the input requires direct interaction with the original feature pairs produced by the Siamese network.
Our best configuration aligns generator features with the separability objective and aligns Siamese features with the reconstruction loss in the auxiliary module. This combination provides a total improvement of 1.8\%.

\begin{table}[t]
  \setlength{\tabcolsep}{3pt}%
  
  \caption{Component-wise improvements in BiCD using the proposed auxiliary objective (Eq.~\ref{eq:auxmoduleIB1}), reported relative to the baseline F1-score (84.7\%).}
  \label{tab: ablation}
  \centering
   \scalebox{0.9}{
  \begin{tabular}{lccc}
    \toprule
    Method 
    & \multicolumn{1}{c}{%
      \begin{tabular}[c]{@{}c@{}}Formula \\ of \(\varPsi\)\end{tabular}
      }
    & \multicolumn{1}{c}{%
      \begin{tabular}[c]{@{}c@{}}Location \\ of Aux\end{tabular}
      } 
    & \multicolumn{1}{c}{%
      \begin{tabular}[c]{@{}c@{}}F1-score \\ (\%)\end{tabular}} \\
    \midrule
    Baseline      
      & \ding{55}  
      & \ding{55}  
      & 84.7   \\
    + BiCD        
      & Eq.~\ref{eq:auxmoduleIB1}  
      & backbone  
      & 84.8   \\
    + BiCD        
      & $I(X,Z(\theta, \eta))$     
      & backbone  
      & 85.2   \\
    + BiCD        
      & Eq.~\ref{eq:auxmoduleIB1}  
      & \begin{tabular}[c]{@{}c@{}}1bit change \\ [-2pt] generator\end{tabular}
      & 85.8   \\
    \textbf{+ BiCD (Ours)}
      & \begin{tabular}[c]{@{}c@{}}
          $\bm{I\!\left( X,\!Z\!\left( \theta ,\!\eta \right) \right)}$\\
          $\textbf{+}$\\
          \textbf{Eq.~\ref{eq:auxmoduleIB1} }\\
        \end{tabular}    
      & \begin{tabular}[c]{@{}c@{}}
          \textbf{backbone}\\
          [-2pt]$\textbf{+}$\\
          [-2pt]\textbf{1bit change}\\
          [-2pt]\textbf{generator}
        \end{tabular}  
      & \textbf{86.5} \\
    \bottomrule
  \end{tabular}}
\end{table}

\textbf{Latency result on edge devices.}
We measure hardware latency by exporting our method within 1-bit C-3PO from PyTorch to ONNX~\cite{onnx} format and deploying it on an ARM Cortex-A76 edge device via the Bolt toolkit~\cite{bolt}. Results (Table~\ref{tab: latency}) show the 1-bit variant achieves 2.5× lower latency than full-precision baselines, while BiCD adds no inference overhead. 
We anticipate further acceleration potential through dedicated hardware accelerators optimized for binarized operations.
\begin{table}[t]
  \setlength{\tabcolsep}{4pt}
  
  \caption{Comparison of latency in deployment on edge device. The input size is 256x256. }
  \label{tab: latency}
  \centering
   \scalebox{0.9}{
  \begin{tabular}{lcccc}
    \toprule
    Method & Bits & \multicolumn{1}{c}{\begin{tabular}[c]{@{}c@{}}Params.\\(M)\end{tabular}} & \multicolumn{1}{c}{\begin{tabular}[c]{@{}c@{}}OPs\\(G)\end{tabular}} & \multicolumn{1}{c}{\begin{tabular}[c]{@{}c@{}}Latency\\(ms)\end{tabular}} \\
    \midrule
    Real-valued C-3PO & 32 & 40.5 & 222.0 & 392.8 \\
    1bit C-3PO        & 1  & 2.1  & 6.6   & 158.4 \\
    1bit C-3PO + BiCD & 1  & 2.1  & 6.6   & 158.4 \\
    \bottomrule
  \end{tabular}}
\end{table}

\section{Conclusion}
\label{sec:conclusion}


We propose BiCD, the first binary neural network for change detection, applying the Information Bottleneck principle to preserve essential information and enhance feature separability. A training-exclusive auxiliary module enhances discernment between meaningful ``interest changes" and irrelevant ``noise changes" without inference costs.  Extensive evaluations on street-view and remote sensing datasets validate BiCD's state-of-the-art performance.

\section*{Acknowledgements}
\label{sec:acknowledgement}
This work was supported by Fundo para o Desenvolvimento das Ciências e da Tecnologia of Macau (FDCT) with Reference No. 0067/2023/AFJ, No. 0117/2024/RIB2, and 0067/2024/ITP2.

{\small
\bibliographystyle{ieeenat_fullname}
\bibliography{11_references}
}

\end{document}